\documentclass[10pt,twocolumn,letterpaper]{article}

\usepackage[final,algorithms]{wacv}      % To produce the REVIEW version for the algorithms track
\usepackage{wacv}              % To produce the CAMERA-READY version
\usepackage{graphicx}
\usepackage{amsmath}
\DeclareMathOperator*{\argmin}{arg\,min}
\usepackage{amssymb}
\usepackage{booktabs}
\usepackage[ruled]{algorithm2e}
\usepackage{multirow}
\usepackage[accsupp]{axessibility} % Improves PDF readability for those with disabilities.

\usepackage[pagebackref,breaklinks,colorlinks]{hyperref}

\usepackage[capitalize]{cleveref}
\crefname{section}{Sec.}{Secs.}
\Crefname{section}{Sec.}{Secs.}
\Crefname{table}{Table}{Tables}
\crefname{table}{Table}{Tables}

\def\wacvPaperID{2463} % *** Enter the WACV Paper ID here
\def\confName{WACV}
\def\confYear{2025}

\begin{document}

\title{
Inverting the Generation Process of Denoising Diffusion Implicit Models: Empirical Evaluation and a Novel Method
}

% \color{red} Evaluating Methods for Predicting Initial Latent Variables from Images\\
% in Denoising Diffusion Implicit Models}

% Improved Inversion Process for Denoising Diffusion Implicit Models}

\author{Yan Zeng$^1$
~~~~Masanori Suganuma$^{1,2}$
~~~~Takayuki Okatani$^{1,2}$\\
$^1$Graduate School of Information Sciences, Tohoku University ~~~~ $^2$RIKEN Center for AIP \\
{\tt\small \{yan,suganuma,okatani\}@vision.is.tohoku.ac.jp}
}

\maketitle

%%%%%%%%% ABSTRACT
\begin{abstract}
This paper studies the problem of inverting the DDIM image generation process to recover latent variables, particularly the initial noise map, from a generated image. Existing methods often struggle with accuracy in this task. We propose a novel hybrid approach that combines direct inversion via gradient descent for the first step, followed by a fixed-point method for subsequent steps. Empirical evaluations across three datasets demonstrate that our method significantly improves the prediction of initial latent variables while achieving superior reconstruction accuracy. Additionally, we introduce a new evaluation, called the self-interpolation test, which assesses the quality of images generated from interpolated points between the true and predicted latent maps, offering deeper insights into performance. Our results reveal that while existing methods perform reasonably well in reconstruction, they consistently fail to accurately predict the initial latent variables, resulting in poor performance on the self-interpolation test. In contrast, our method outperforms all others across all metrics, providing valuable insights into diffusion models and enhancing their applications in image generation and editing.

\end{abstract}

%%%%%%%%% BODY TEXT
\section{Introduction}
\label{sec:intro}

Recent advancements in image generation using diffusion models have been substantial. In a diffusion process, noise is incrementally added to an image at each step, $x_{t-1} \rightarrow x_t$. When this process is applied to a clean image, such as a natural image $x_0$, and repeated over many steps ($t=1,2,\ldots,T$), it ultimately transforms the image into a fully noisy map $x_T$ as $T$ becomes large. Diffusion models reverse this process, gradually recovering $x_0$ from $x_T$. Typically, denoising from $x_t$ to $x_{t-1}$ is carried out by a deep network, often a U-Net. This network is trained on a set of target images $\{x\}$, allowing it to generate images that follow the distribution $p(x)$ of that set. Using the trained U-Net (or a similar model), an image $x_0 \sim p(x)$ can be generated from any sampled noise image $x_T$.

The noisy images $x_t$ (for $t \neq 0$) in the intermediate steps of the denoising process affect the final generated image $x_0$ and can be considered its latent variables. Understanding the relationship between these latent variables and the generated image is essential, not only for applications such as image editing \cite{meng2022sdedit} but also for a deeper understanding of the behavior of diffusion models. Despite this, there is currently no method to accurately infer the latent variables $x_t$ from $x_0$, whether $x_0$ is a generated image or a natural image.

\begin{figure}[t]
\centering
\includegraphics[width=0.80\columnwidth]{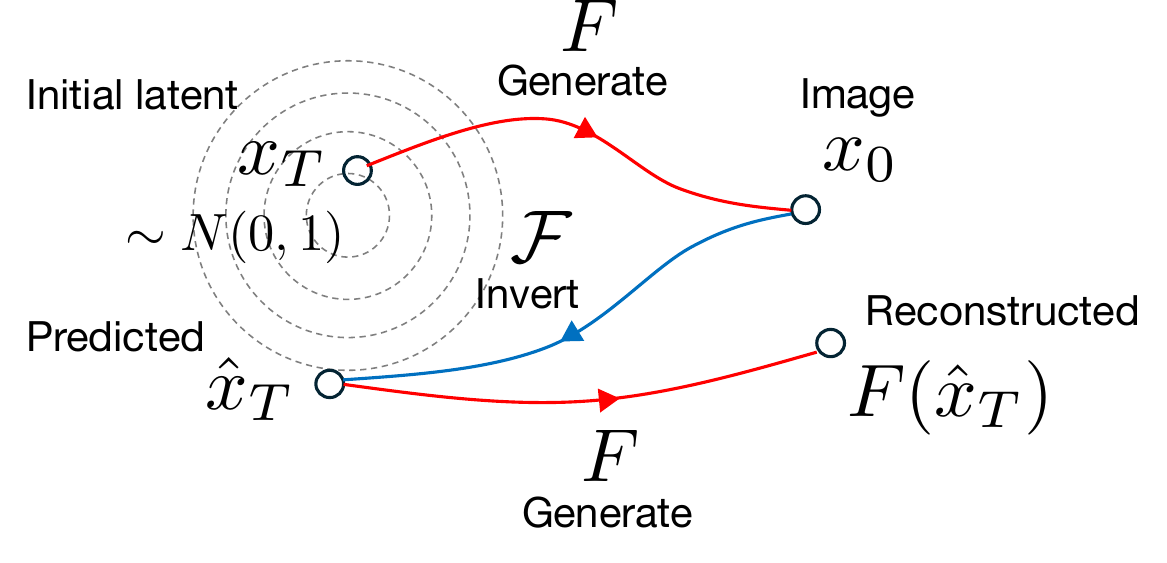}
\caption{DDIM generation process ($F:x_T\rightarrow x_0$) and its inversion ($\mathcal{F}:x_0\rightarrow \hat{x}_T$).}
\label{fig:diagram}
\end{figure}

In DDPM \cite{ho2020denoising}, the generation process is stochastic, so the latent variables cannot be uniquely identified. However, in DDIM \cite{song2021denoising}, where the generation is deterministic, the latent variables are directly tied to the final generated image, suggesting that they could be uniquely determined, as shown in Fig.~\ref{fig:diagram}. As discussed later, several studies focused on image editing have explored and applied this approach, yielding promising results \cite{couairon2023diffedit}\cite{mokady2023null}\cite{parmar2023zero}.

Building on the above, this paper addresses the problem of inverting the DDIM image generation process to recover its latent variables, particularly the initial latent $x_T$, which is expected to be a complete noise map, from a generated image $x_0$ or even an image that was not generated by DDIM.

Extensive research has been conducted on reversing the image generation process in diffusion models. Early efforts focused on methods that added noise to replicate the standard diffusion process, as seen in works such as SDEdit \cite{meng2022sdedit} and Blended Diffusion \cite{avrahami2022blended}. Later, a technique known as DDIM inversion \cite{song2021denoising} was developed by drawing parallels between the DDIM process and the Euler method for solving ordinary differential equations, enabling the reverse computation. This approach has since been applied in various image editing techniques, including DDIB \cite{su2022dual}, DiffusionCLIP \cite{kim2022diffusionclip}, and Prompt-to-Prompt \cite{hertz2023prompttoprompt}.

However, DDIM inversion relies on a rough approximation, and it is known that the accuracy remains low unless the number of generation steps, $T$, is sufficiently large. To overcome this limitation, several extensions and improvements have been introduced. Parmar et al. \cite{parmar2023zero} noted that the noise estimated by the U-Net in DDIM inversion deviates from the expected Gaussian distribution. They proposed a method to refine the predicted latent variables, making them more Gaussian by applying techniques such as autocorrelation loss and KL loss to adjust the latent variables in a way that minimizes these discrepancies. In contrast, Pan et al. \cite{pan2023effective} and Meiri et al. \cite{meiri2023fixed} introduced an alternative approach that iteratively estimates the latent variables using a fixed-point method derived from the one-step DDIM computation. Additionally, hybrid methods that combine elements from both of these approaches have also been proposed \cite{garibi2024renoise}.

In this study, we empirically evaluate existing methods under the same experimental settings and introduce a novel approach that demonstrates superior performance. Our method, though intuitive, has not been reported in previous literature: it directly inverts one-step DDIM generation through gradient descent optimization. More specifically, we employ this direct inversion technique only in the first step of the inversion process and apply the existing fixed-point method for all subsequent steps, creating a hybrid approach. Our experiments show that this method outperforms all existing techniques.

In this paper, unlike the aforementioned studies, we focus on the simplest DDIM image generation setting, i.e., unconditional DDIM in the image space. This is motivated by our primary interest in determining how to achieve high-precision DDIM inversion, starting with the most fundamental form of DDIM. In contrast, many previous studies were concerned with image editing in text-to-image synthesis, targeting conditional DDIM in latent diffusion models (LDMs) \cite{rombach2022high}. Instead of operating in the image space, these studies first map the images into a smaller latent variable space using a VAE, where conditional DDIM, guided by text input, is applied for inversion.

In the experiments, we evaluated both the existing methods and the proposed method using three datasets. We used three evaluation metrics: the accuracy of the estimated latent variable $\hat{x}_T$, the accuracy of the image generated (i.e., reconstructed) from it using DDIM, and the accuracy of the image generated from the interpolated points between latent variables. The last metric, based on interpolation, involves generating an image starting from an interpolated point between the latent variable $x_T$ of a given image and its estimate $\hat{x}_T$, and then assessing the accuracy and quality of the resulting image. This approach leverages the property that proper interpolation between legitimate latent variables should itself result in a valid latent variable. While a similar idea has been used to test generative models before, interpolating between the latent variables of the same image is novel. Although simple because it does not require two different images, this metric is highly effective in quantifying the performance differences between inversion methods.

Several insights were gained. First, methods based on DDIM inversion have low accuracy in reconstruction but perform well in interpolation. On the other hand, methods based on the fixed-point method excel in reconstruction but perform poorly in interpolation. Most importantly, the proposed method outperforms these existing methods in both reconstruction and interpolation.

%------------------------------------------------------------------------
\section{Background}
\label{sec:background}

\subsection{Denoising Diffusion Model}

Diffusion models begin with a noise map sampled from a Gaussian distribution and iteratively reduce the noise to generate an image. This process typically involves a deep network, most commonly a U-Net. At each step, the U-Net receives a noisy image along with the current timestep as input and predicts the noise. The predicted noise is then used to update the noisy image, progressively reducing the noise. In the original Denoising Diffusion Probabilistic Model (DDPM), the update formula is similar to Langevin dynamics, as follows:
\begin{equation*}
    x_{t-1}=\frac{1}{\sqrt{\alpha_t}}
    \left(x_t-\frac{1-\alpha_t}{\sqrt{1-\bar{\alpha}_t}}\epsilon_\theta(x_t,t)\right)+\sigma_t{z},
\end{equation*}
where $x_{t-1}$ and $x_t$ represent the noisy images at timesteps $t-1$ and $t$, respectively, $\alpha_t$, $\bar{\alpha}_t$, and $\sigma_t$ are scheduler-determined coefficients, $\epsilon_\theta(x_t, t)$ is the noise predicted by the U-Net, and $z$ is noise sampled from a Gaussian distribution, introducing randomness at each step.

\subsection{Denoising Diffusion Implicit Model}

Because the generation process in diffusion models follows a Markov chain, DDPM is significantly slower than other generative models. In \cite{nichol2021improved}, the authors introduced a strided sampling scheduler to reduce the number of generation steps. Building on this, \cite{song2021denoising} further modified the denoising equation, leading to the denoising diffusion implicit model (DDIM) with the following update formula: 
\begin{equation}\label{eq:ddim_o}
    x_{t-1}=\frac{x_t}{\sqrt{\alpha_t}}+(\sqrt{1-\bar{\alpha}_{t-1}}-\frac{\sqrt{1-\bar{\alpha}_t}}{\sqrt{\alpha_t}})\epsilon_\theta(x_t,t).
\end{equation}
DDIM significantly reduces the number of steps while maintaining high-quality image generation.

Unlike DDPM, deterministic DDIM removes the stochastic noise term from the inference process. In addition to speeding up the generation process, a key advantage of DDIM is its deterministic nature. Given an initial latent variable, each denoising step with DDIM produces a deterministic result. Iterating through these steps yields a fully deterministic final image, meaning that when using DDIM for inference, the generated image is uniquely determined by the initial latent.

%------------------------------------------------------------------------
\section{Methods for Predicting Initial Latent}

In this section, we summarize several methods for reversing the image generation process in DDIM, specifically predicting the initial latent 
$x_T$ from a generated image $x_0$. 

\subsection{Basic Methods}
\label{sec:basic_methods}

We first present several basic methods for estimating 
$x_T$ from $x_0$. Each method reverses an unconditional generation process, meaning 
$x_0$ is generated from $x_T$ without any additional input.

\subsubsection{DDIM Inversion}

In the DDIM paper \cite{song2021denoising}, the authors present a method for reversing the generation process, called {\em DDIM inversion}. This approach leverages the similarity between DDIM's iterative computations and Euler's method for solving ordinary differential equations. The authors indicate that, with a sufficiently large number of generation steps, the process can be effectively reversed.

The DDIM iteration (\ref{eq:ddim_o}) can be simplified as 
\begin{equation}\label{eq:ddim_gen}
    x_{t-1}=m_t{x_t}+n_t{\epsilon_\theta(x_t,t)},
\end{equation}
where $m_t$ and $n_t$ are the coefficients determined by $\alpha_t$ and $\bar{\alpha}_t$, which are given by the scheduler. By rearranging the equation, we obtain
\begin{equation}\label{eq:ddim_inv}
    x_t=\frac{x_{t-1}-n_t\epsilon_\theta(x_t,t)}{m_t},
\end{equation}
which serves as a basic formula for DDIM inversion. 
As the value of $x_t$ inside $\epsilon_\theta$ on the right-hand side is unknown, DDIM inversion introduces an approximation
\begin{equation}\label{eq:lin_approx_assump}
    \epsilon_\theta(x_t,t)\approx\epsilon_\theta(x_{t-1},t)
\end{equation}
for all inversion steps from $t=1$ to $T$. This approximation will be more accurate if the number of generation steps is larger. 

However, we have empirically observed that this approximation becomes less accurate for timesteps approaching 0. While the assumption holds reasonably well during the later stages of DDIM inversion, the error becomes significant during the first inversion steps, where $t$ is close to 0. This issue persists even when the number of generation steps is increased to a large value, such as $T=1,000$. Consequently, the inverse trajectory deviates from the original generation trajectory, leading to substantial errors in the predicted initial latent.

\subsubsection{Fixed-point Iteration}

In \cite{pan2023effective}, the authors proposed treating the reversal process as a fixed-point problem. This approach was also adopted in \cite{garibi2024renoise} and can be summarized as follows.

Let $h(\xi)$ be the right-hand side of (\ref{eq:ddim_inv}) as follows:
\begin{equation}
    h(\xi)\equiv\frac{x_{t-1}-n_t\epsilon_\theta(\xi,t)}{m_t}.
\end{equation}
A fixed-point iteration aims to find a solution $\hat{x}_t$ to the equation $\hat{x}_t=h(\hat{x}_t)$ by updating an estimate $\hat{x}_t^n$ iteratively as
\begin{equation}
    \hat{x}_t^{n+1}=h(\hat{x}_t^n)
\end{equation}
for $n=1,2,\ldots$, starting from an initial guess $\hat{x}_t^1$. This approach can also be viewed as solving the following optimization problem: 
\begin{equation}
    \hat{x}_t = \argmin_{\xi}\lVert \xi-h(\xi)\rVert_2
\end{equation}

\subsubsection{Direct Optimization by Gradient Descent}

A single DDIM step calculates $x_{t-1}$ from $x_t$ according to (\ref{eq:ddim_gen}). Letting the right-hand side of this equation be $f(\xi) \equiv m_t\xi+n_t\epsilon_\theta(\xi,t)$, we can write it as $x_{t-1} = f(x_t)$. One step of the reverse process is to predict $x_t$ from $x_{t-1}$ by inverting this equation. 

This can be formulated as the following optimization problem:
\begin{equation}
    \hat{x}_t=\argmin_{\xi}\lVert x_{t-1}-f(\xi)\rVert_2.
\end{equation}
We consider solving it using gradient descent (GD) and refer to this method as {\em direct optimization} in what follows. The function $f$ is a linear combination of $x_t$ and the output obtained by feeding $x_t$ and $t$ into a DNN, such as a U-Net. Essentially, this is equivalent to determining the input of the DNN from its output \cite{mahendran2015understanding}. 

If $f$ is an injective function, then the solution should be unique. However, proving that $f$ is injective is beyond the scope of this study. We only expect that $\hat{x}_t$ approximates $x_t$. 

Although this method is simple, it has not been addressed in previous research. Later, we will discuss how this method performs during inversion. In brief, experiments indicate that it {\em only works at the first step, but at that step, it significantly outperforms other methods.}

\subsection{Advanced Methods}

In prior research on image editing methods, more advanced inversion techniques, building on the basic methods mentioned earlier, have been proposed. This section summarizes these studies. However, they focus primarily on text-to-image synthesis models, which are often in high demand for practical applications, such as Stable Diffusion. Consequently, the image generation models in these studies differ from the DDIM model considered in this paper. First, they apply DDIM in the latent variable space mapped by a VAE, specifically in a latent diffusion model (LDM) \cite{rombach2022high}. Thus, the following studies focus on DDIM inversion in this latent space. Second, the generation is conditional, as their targets are text-to-image synthesis models. As DDIM functions similarly regardless of whether it operates in the image space or latent variable space and whether it is conditional or unconditional, we port those inversion methods to unconditional, image-space DDIM. However, their effectiveness may diverge from the conclusions drawn in their respective papers, which will also be discussed below.

\subsubsection{AIDI}

AIDI (Accelerated Iterative Diffusion Inversion) \cite{pan2023effective} is a method designed to accurately and efficiently compute the latent representation of an input image, with the goal of enhancing the usability and quality of image editing. It essentially utilizes the aforementioned fixed-point method as its core for the inversion process. Instead of a simple fixed-point iteration, where the current result is used as the input for the next step, AIDI employs a technique akin to Anderson acceleration to improve the trade-off between accuracy and efficiency. Specifically, a weighted average of the results from several previous iterations is used as the input for the next step.

There are two versions of AIDI, -A and -E. The former involves averaging the results of multiple iterations, with the weights learned through optimization, while the latter averages the two most recent iterations with equal weights. The results of these two approaches are similar, and the latter is used in their experiments. In our unconditional setting, we found that the iterations converge quickly, and thus averaging the last two iterations yields the same result as directly using the previous iteration's output. Therefore, in our reproduction of AIDI and in our method, we opted to use the naive iterative approach.

\subsubsection{pix2pix-zero}

pix2pix-zero \cite{parmar2023zero} is an image editing method designed for text-to-image synthesis models based on an LDM. It introduces a novel approach to improving the inversion of the DDIM generation process. The authors observe that the noise predicted by the U-Net at each step of DDIM inversion often fails to exhibit the statistical properties of Gaussian white noise. They argue that this degrades the inversion quality, leading to a decline in editing performance when using the predicted latent. To address this issue, they propose a technique called autocorrelation regularization. This approach employs a loss function that captures the properties expected of true Gaussian noise. By minimizing this loss, the predicted noise is adjusted through gradient descent at each step of DDIM (with five iterations used in their experiments). The loss function has two components: pairwise regularization and KL divergence regularization. The former reduces the correlation between pixels in the predicted noise map, ensuring pixel independence. The latter ensures that the noise values for each pixel conform to a Gaussian distribution with zero mean and unit variance, a key condition for Gaussian noise.

\subsubsection{ReNoise}

ReNoise \cite{garibi2024renoise} is a method based on the fixed-point approach, incorporating the correction introduced in pix2pix-zero that adjusts the predicted noise to resemble a Gaussian distribution. It demonstrates superior performance in both image reconstruction accuracy and processing speed compared to existing inversion methods. It is particularly claimed to be effective for diffusion models trained with a small number of denoising steps. Additionally, an extended method for DDPM inversion, referred to as ``Noise-Correction'' in the paper, has been proposed, but it is omitted here because it is not relevant to our study.

\subsubsection{Proposed Hybrid Inversion Method}

Among the basic methods discussed in \cref{sec:basic_methods}, direct optimization, to the best of our knowledge, has not been explored in existing research. Through preliminary experiments, we made the following findings.

First, this method only works effectively in the first step of the inversion, i.e., $x_0 \rightarrow x_1$. For $t = 2, 3, \ldots$, it produces highly inaccurate results for unknown reasons. Second, direct optimization tends to perform significantly better during the first inversion step than the other basic methods. 

Although we lack a formal explanation for this phenomenon, a plausible reason is that the input $x_{t-1}$ to the inversion $x_{t-1} \rightarrow x_t$ always contains some noise, except at $t=1$; $x_0$ is the generated noise-free image, while all other $x_t$ values ($t \geq 1$) contain some level of noise. The existence of noise in the target $x_{t-1}$ may make the optimization $\|x_{t-1}-f(\xi)\|_2$ more difficult. 

Building on this experimental finding, we propose a method that applies direct optimization at the first inversion step ($x_0 \rightarrow x_1$) and the fixed-point method at the subsequent steps ($t=2, \ldots, T$). The method is summarized in \cref{pseudo}.

\begin{algorithm}
\DontPrintSemicolon
    \KwIn{an image $x_0$}
    \KwOut{a predicted initial latent $\hat{x}_T$}
    $\hat{x}_1\leftarrow\argmin_{\xi}\lVert x_0-f(\xi)\rVert_2$\;
    \For{$t=2,\ldots,T$}{
    $\hat{x}_t^0\leftarrow\hat{x}_{t-1}$\;
    \For{$i=0,\ldots,I$}{
    $\hat{x}_t^{i+1}\leftarrow h(\hat{x}_t^i)$\;}
    $\hat{x}_t\leftarrow\hat{x}_t^{I+1}$\;}
    \Return$\hat{x}_T$\;
\caption{Proposed Inversion Method}\label{pseudo}
\end{algorithm}

\section{Experiments}

We experimentally compare the above methods using three datasets: CelebA \cite{liu2015deep}, LSUN Bedroom, and LSUN Church \cite{yu2015lsun}. To facilitate reproducibility, we utilize pre-trained DDIM models provided by Google on Hugging Face\footnote{\url{https://huggingface.co/google/ddpm-celebahq-256}, \url{https://huggingface.co/google/ddpm-bedroom-256}, \url{https://huggingface.co/google/ddpm-church-256}}, each specifically trained on one of the datasets.

\subsection{Experimental Settings}

We conduct the experiments as follows. First, we sample an initial latent variable $x_T$ from a Gaussian distribution $\mathcal{N}(0, I)$, which serves as the starting point. Using a pre-trained DDIM model, we generate an image $x_0$ from $x_T$, denoted as $x_0 \equiv F(x_T)$, where $F(\cdot)$ represents the DDIM model. Next, we apply each inversion method $\mathcal{F}$ to $x_0$ to estimate the initial latent variable, denoted as $\hat{x}_T \equiv \mathcal{F}(x_0)$. This process is repeated by sampling new latent variables $x_T$, and the results are averaged.

\begin{figure}[t]
\centering
\includegraphics[width=0.7\columnwidth]{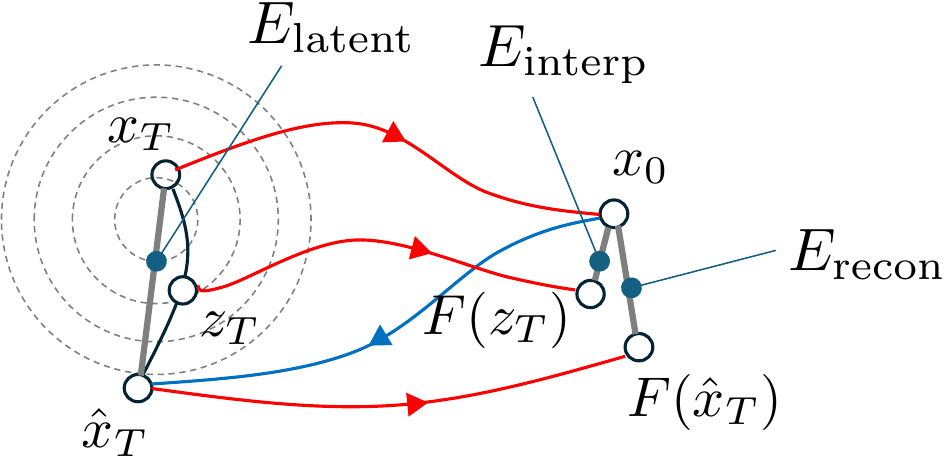}
\caption{Illustration of the three employed metrics.}
\label{fig:metrics}
\end{figure}

To evaluate the accuracy of each inversion method, we use three metrics; see Fig.~\ref{fig:metrics}. For two vectors $a,b\in\mathbb{R}^d$, we define their mean squared error (MSE) as $\operatorname{MSE}(a,b)\equiv \frac{1}{d}\lVert a-b\rVert_2^2$. The first metric is the MSE of the estimated initial latent $\hat{x}_T$, calculated as
\begin{equation}
    E_\mathrm{latent}\equiv \operatorname{MSE}(\hat{x}_T, x_T).
\end{equation}
While this is a straightforward metric, its interpretability is limited. Moreover, it does not reflect how closely the estimates $\hat{x}_T$ adhere to a Gaussian distribution.

The second metric evaluates the accuracy of the reconstructed image from $\hat{x}_T$. Specifically, using the same DDIM model, we reconstruct the original image, i.e., generate an image as $F(\hat{x}_T)$, and measure the difference between the original $x_0$ and its reconstruction $F(\hat{x}_T)$ using MSE, defined as 
\begin{equation}
    E_\mathrm{recon}\equiv \operatorname{MSE}(F(\hat{x}_T), x_0).
    % =\lVert F(\hat{x}_T) - F(x_T)\rVert_2. 
\end{equation}

The third evaluation method involves image interpolation. Given two latent variables, $x_T$ and $y_T$, we compute their interpolation as $z_{T,\alpha} \equiv \mathrm{slerp}(x_T, y_T; \alpha)$, where $\mathrm{slerp}$ denotes spherical linear interpolation, and $\alpha \in [0, 1]$ is the interpolation parameter. We use $\mathrm{slerp}$ for its appropriateness in this context \cite{white2016sampling}. The model then generates an image as $F(z_{T,\alpha})$, and we assess its quality.

Image interpolation is commonly used to evaluate generative models. For example, generative models that merely memorize training samples can be identified because the interpolated images will either be of low quality or exhibit clear artifacts. In contrast, if the model learns a smooth and meaningful latent space, the interpolated images should maintain high quality and reflect intermediate semantics between the two original samples.

To enable stable, quantitative evaluation, we propose a simple method called \emph{self-interpolation}. Instead of interpolating the latent variables of two different images, we interpolate between the true initial latent variable, $x_T$, and its prediction, $\hat{x}_T$, for the same image. The DDIM model generates an image from their interpolation, $z_{T,\alpha} \equiv \mathrm{slerp}(x_T, \hat{x}_T; \alpha)$. If the difference between $x_T$ and $\hat{x}_T$ is small, the generated image $F(z_{T,\alpha})$ should closely resemble the original image, $x_0$. However, if there is a significant gap, $F(z_{T,\alpha})$ will deviate from $x_0$. Setting $\alpha=0.5$, we define the following metric:
\begin{equation}
    E_\mathrm{interp}\equiv \operatorname{MSE}(F(z_{T,0.5}), x_0).
\end{equation}
% {\color{red} \bf [What $\alpha$ is used when calculating $E_\mathrm{interp}$?]}{\color{green} [it is set to 0.5, which is the middle of the latents]}
At first glance, this metric may seem easily minimized by any model and therefore appear trivial, but as we will show later, existing methods actually fail dramatically.

\begin{table*}[t]
    \centering
     \caption{
    Comparison of different inversion methods in terms of three evaluation metrics: the MSEs of the estimated initial latent images, the reconstructed images, and the images generated through self-interpolation.}\small
    
    \begin{tabular}{c c c c | c c c | c c c} 
    \hline
    \multirow{2}{*}{}&
        \multicolumn{3}{c|}{CelebA} &
        \multicolumn{3}{c|}{LSUN Bedroom} &
        \multicolumn{3}{c}{LSUN Church}\\
    & $E_\mathrm{latent}$ & $E_\mathrm{recon}$ & $E_\mathrm{interp}$ & $E_\mathrm{latent}$ & $E_\mathrm{recon}$ & $E_\mathrm{interp}$ & $E_\mathrm{latent}$ & $E_\mathrm{recon}$ & $E_\mathrm{interp}$\\
    \hline
    DDIM inv.{\color{black} \cite{song2021denoising}} & 0.7951 & $2.9426\mathrm{e}{-02}$ & 0.4504 & 0.5998 & $1.9770\mathrm{e}{-02}$ & 0.3688 & 0.7075 & $3.1042\mathrm{e}{-02}$ & 0.4013\\
    \hline
    pix2pix-zero\cite{parmar2023zero} & 0.7958 & $3.2506\mathrm{e}{-02}$ & 0.4538 & 0.5972 & $2.3419\mathrm{e}{-02}$ & 0.3800 & 0.7073 & $3.1916\mathrm{e}{-02}$ & 0.4066\\
    \hline
    AIDI\cite{pan2023effective} & 0.8789 & $1.9479\mathrm{e}{-05}$ & 1.3036 & 0.1506 & $1.2454\mathrm{e}{-04}$ & 0.1178 & 0.3799 & $1.8451\mathrm{e}{-04}$ & 0.3351\\
    \hline
    ReNoise\cite{garibi2024renoise} & 0.8787 & $2.1284\mathrm{e}{-05}$ & 1.3020 & 0.1507 & $1.8123\mathrm{e}{-04}$ & 0.1175 & 0.3797 & $2.8800\mathrm{e}{-04}$ & 0.3344\\
    \hline
    Ours & 0.2388 & $1.8493\mathrm{e}{-05}$ & 0.1389 & 0.0920 & $9.5469\mathrm{e}{-05}$ & 0.0610 & 0.2437 & $1.8795\mathrm{e}{-04}$ & 0.1824\\
    \hline
    \end{tabular}
    \label{tab:comp}
\end{table*}

\begin{figure*}[t]
    \centering
        \begin{subfigure}{0.30\linewidth}
        \includegraphics[width=\linewidth]{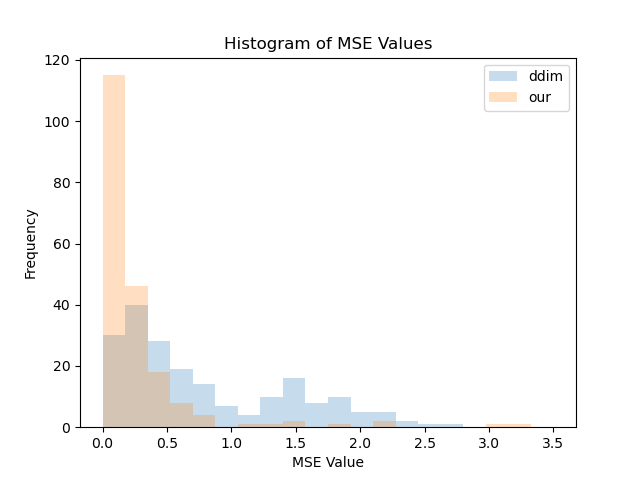}
        \caption{}
        \end{subfigure}
        \begin{subfigure}{0.30\linewidth}
        \includegraphics[width=\linewidth]{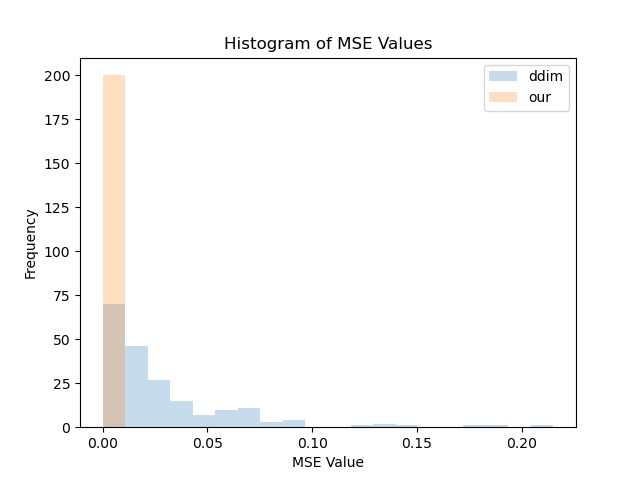}
        \caption{}
        \end{subfigure}
        \begin{subfigure}{0.30\linewidth}
        \includegraphics[width=\linewidth]{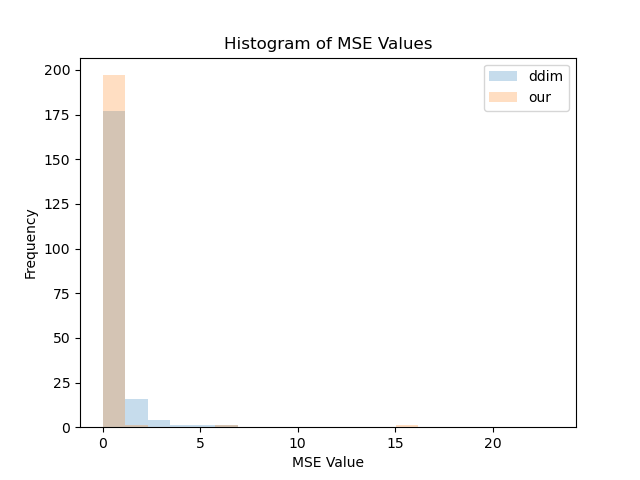}
        \caption{}
        \end{subfigure}
    \vfill
        \begin{subfigure}{0.30\linewidth}
        \includegraphics[width=\linewidth]{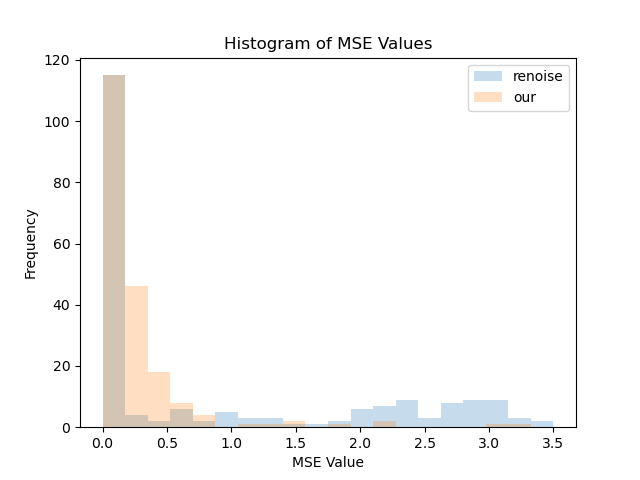}
        \caption{}
        \end{subfigure}
        \begin{subfigure}{0.30\linewidth}
        \includegraphics[width=\linewidth]{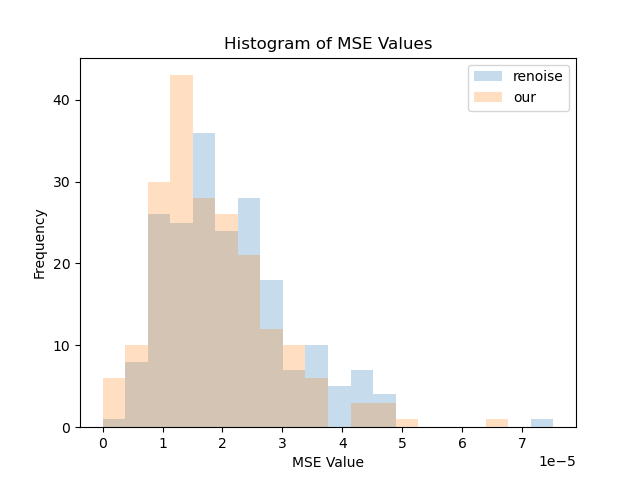}
        \caption{}
        \end{subfigure}
        \begin{subfigure}{0.30\linewidth}
        \includegraphics[width=\linewidth]{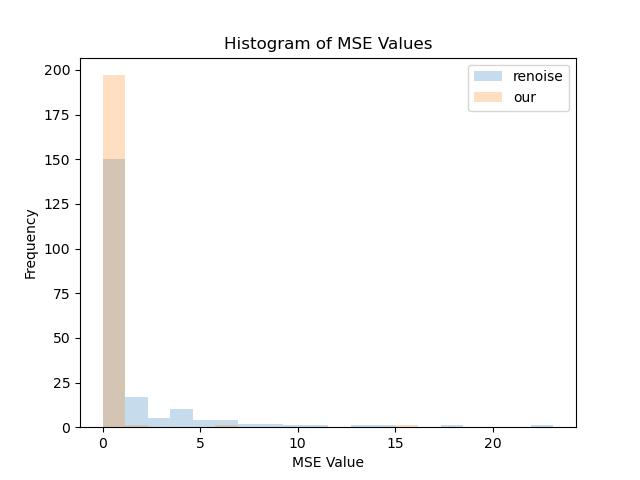}
        \caption{}
        \end{subfigure}        
    \caption{These histograms are derived from the results in \cref{tab:comp}. (a)--(c) Compared with DDIM, our method achieves better overall performance in latent prediction, image reconstruction, and self-interpolation. (d) ReNoise performs even worse than DDIM in latent prediction. (e) Both ReNoise and our method demonstrate excellent performance in image reconstruction. (f) However, like DDIM, ReNoise performs poorly in self-interpolation.}
    \label{fig:hist_all}
\end{figure*}

\subsection{Results}

\begin{figure*}[t]
    \centering
    \includegraphics[width=\linewidth]{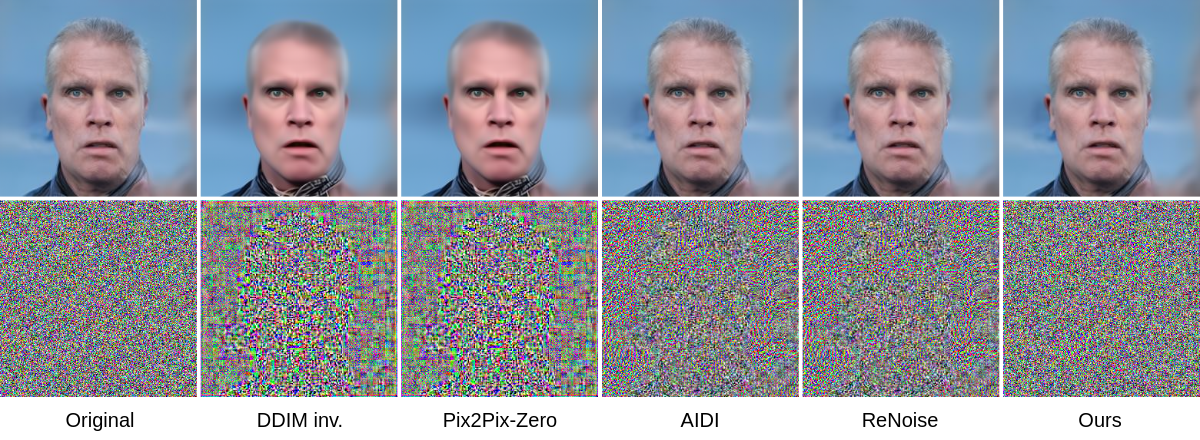}
    \caption{The leftmost column shows the generated image and its initial latent map. The subsequent columns display the predicted latent maps (bottom) and their corresponding reconstructed images (top) for DDIM inversion, pix2pix-zero, AIDI, ReNoise, and our method. Compared to the original image, the reconstructions by DDIM inversion and pix2pix-zero exhibit visible errors, while those produced by the other methods closely resemble the original image. Notably, only our method predicts a latent map without noticeable artifacts.}

   \label{fig:latent_vis}
\end{figure*}

\subsubsection{Quantitative Evaluation}

\cref{tab:comp} presents the results of the three metrics across the different inversion methods. 
The histograms comparing these error metrics for our method against DDIM inversion and ReNoise are displayed in \cref{fig:hist_all}. Several observations can be made from this comparison.

First, the four existing methods can be categorized into two groups: (1) DDIM inversion-based methods, including DDIM inversion and pix2pix-zero, and (2) fixed-point iteration-based methods, including AIDI and ReNoise. The methods within each group produce similar results.

When comparing these two groups, the DDIM inversion-based methods outperform the fixed-point iteration-based methods in terms of latent estimation errors ($E_\mathrm{latent}$) and self-interpolation errors ($E_\mathrm{interp}$). However, this order is reversed when considering reconstruction errors ($E_\mathrm{recon}$). This suggests that methods in the second group, AIDI and ReNoise, tend to produce estimates of $\hat{x}_T$ that are farther from $x_T$, yet still generate images closer to the true image, i.e., $F(\hat{x}_T) \sim x_0$.

The second group's poorer self-interpolation results align with their less accurate estimates of the initial latent. These estimates may not exhibit the characteristics of independent Gaussian noise. As a result, even though these estimates can generate high-quality images, interpolating between latents can lead to poor-quality image generation. To address the deviation from Gaussian noise characteristics, pix2pix-zero and ReNoise employ autocorrelation loss as a constraint to ensure that the estimated initial latents conform to these properties. (Recall that the main methodological difference within each group is the use of autocorrelation loss.) However, the results show that this approach has minimal impact.

Most importantly, our method achieves the lowest errors across all three metrics. It significantly reduces errors in both initial latent estimation and self-interpolation, while also outperforming AIDI and ReNoise in terms of reconstruction errors.

\begin{figure}[tb]
\centering
\includegraphics[width=\columnwidth]{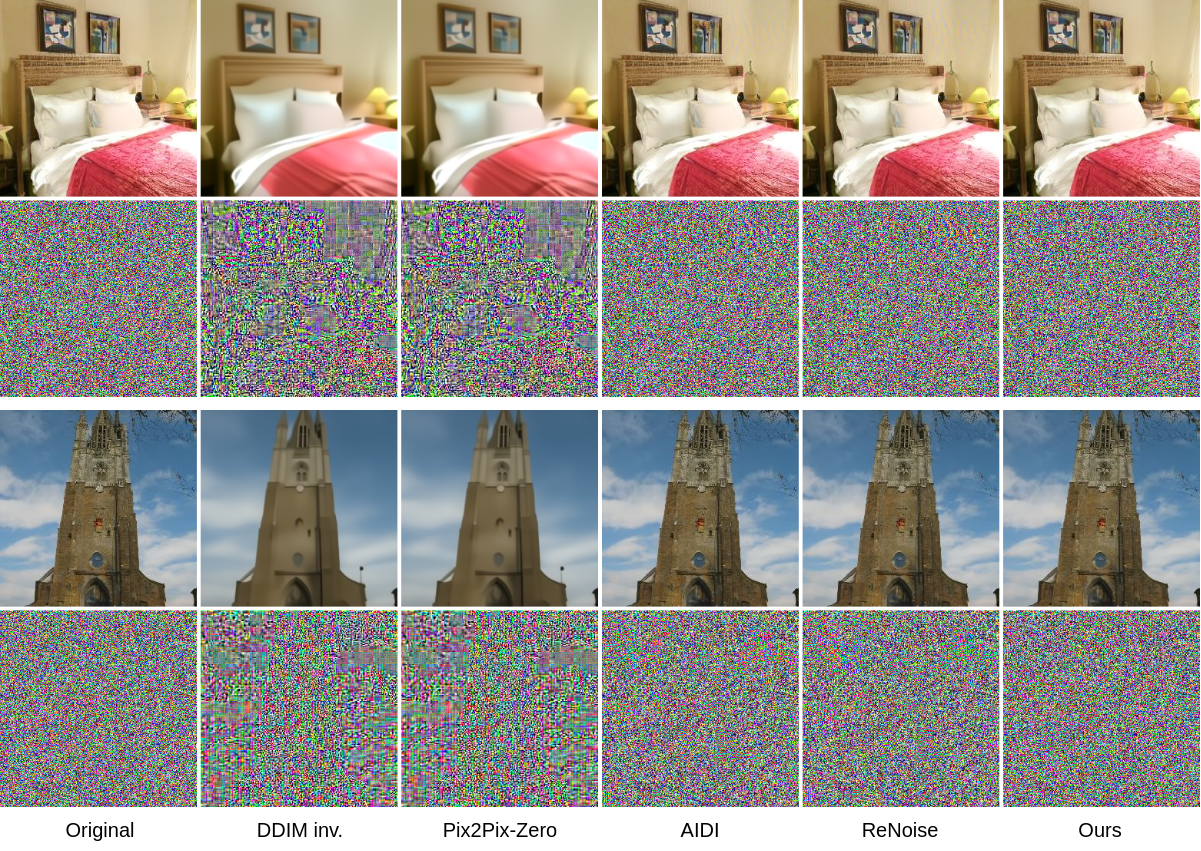}
\caption{Reconstructed images and predicted latent by the compared methods for LSUN Bedroom (top) and Church (bottom), presented in the same format as Fig.~\ref{fig:latent_vis}. }
\label{fig:lsun_vis}
\end{figure}

\subsubsection{Qualitative Comparison}

Figure~\ref{fig:latent_vis} presents examples of the estimated initial latent images, $\hat{x}_T$, produced by the compared methods, along with the corresponding reconstructed images, $F(\hat{x}_T)$. It is clear that the DDIM inversion-based methods (i.e., DDIM inversion and pix2pix-zero) struggle to accurately reconstruct the original image, while the fixed-point iteration-based methods (i.e., AIDI and ReNoise) produce high-quality reconstructions.

However, there is a noticeable difference between the predicted initial latents from the fixed-point iteration-based methods and the true latent. Specifically, the estimates from AIDI and ReNoise resemble the true image locally but exhibit global artifacts, such as a visible outline around the person's head. In contrast, our method produces both high-quality image reconstructions and initial latent estimates without such artifacts. These results are consistent with the previous observations. Similar results for the LSUN Bedroom and Church datasets are shown in Fig.~\ref{fig:lsun_vis}, where the same patterns are evident.

\begin{figure*}[t]
    \centering
    \includegraphics[width=\linewidth]{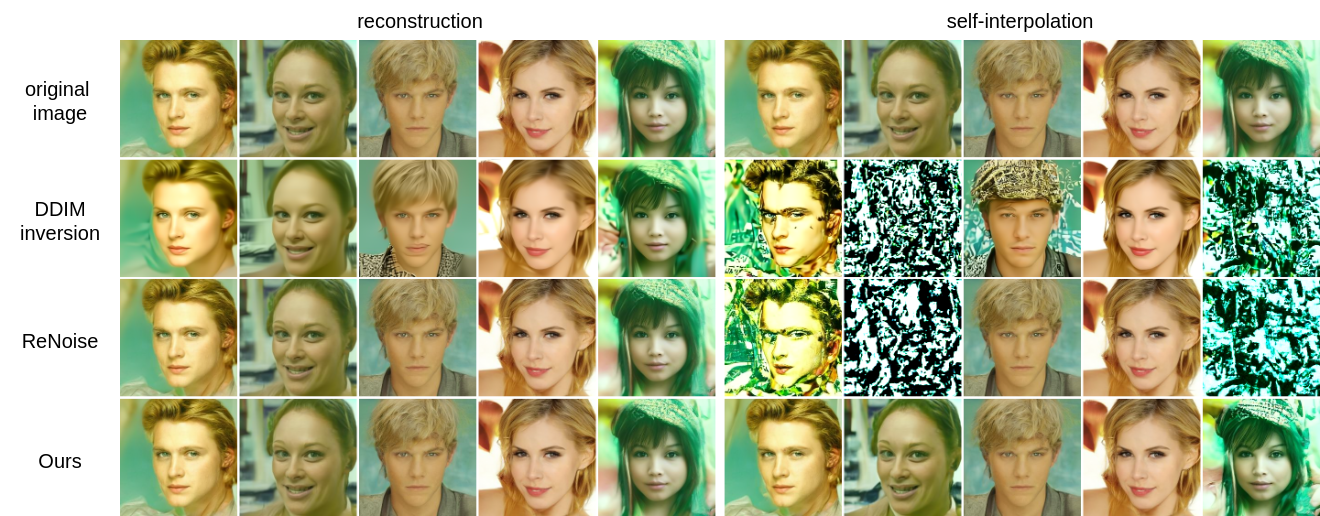}
    \caption{Reconstructed images for standard reconstruction and self-interpolation. Left: Images reconstructed from $\hat{x}_T$. Right (self-interpolation): Images reconstructed from interpolated latent maps $z_{T,\alpha} \equiv \mathrm{slerp}(x_T, \hat{x}_T; 0.5)$. While existing methods perform well in standard reconstruction, they fail in the self-interpolation test.}

   \label{fig:comp_all}
\end{figure*}

Figure~\ref{fig:comp_all} presents several reconstruction results for the standard reconstruction and self-interpolation using different methods. For the standard reconstruction, it can be seen that DDIM inversion produces reasonable results, while both ReNoise and the proposed method provide nearly perfect outcomes. However, in self-interpolation, both DDIM inversion and ReNoise frequently produce corrupted images, with success being very rare. In contrast, the proposed method consistently delivers stable and good results.

%\subsection{Image Reconstruction}

\subsubsection{Computational Cost}

Different methods incur different computational costs. For a simple but practical evaluation, we measure the runtime of each method on the same experimental platform. The experiments were conducted on an RTX A6000 GPU using PyTorch version 2.1.0 and Hugging Face Diffusers version 0.30.0. We set the number of DDIM steps to 50.

DDIM inversion takes approximately 1.5 seconds. pix2pix-zero, using the original settings (i.e., five optimization steps per inversion step), requires about 3.3 seconds. The fixed-point iteration-based methods, AIDI and ReNoise, have runtimes that scale linearly with the number of iterations. With 10 iterations, their runtimes are 15 seconds and 32 seconds, respectively. 

Our method requires additional time in the first step, where we perform 1,000 gradient descent parameter updates, resulting in a total runtime of 76 seconds.

Although the proposed method sacrifices computational efficiency for higher inversion quality, the trade-off is not excessively large. Future work will focus on improving its speed.

\section{Conclusion and Limitation}

This study explored methods for inverting the DDIM image generation process, focusing specifically on recovering latent variables in the simplest form of unconditional DDIM in image space.

We introduced a novel hybrid approach that applies gradient descent in the initial inversion step, followed by the fixed-point method. Through consistent experimental evaluation, we demonstrated significant improvements in the accuracy of latent variable estimation compared to existing methods. Additionally, we proposed a new evaluation metric based on interpolating between the predicted and true latent variables of the same image, providing another effective means to assess inversion techniques.

Our findings indicate that while current DDIM inversion methods achieve good reconstruction results, they face limitations in accurately recovering the initial latent variables. In contrast, the proposed method provides a more robust solution, excelling in both reconstruction and latent variable accuracy. We believe our results will contribute to future advancements in high-precision DDIM inversion. Although this study focused on unconditional DDIM, we plan to extend these insights to the development of more complex models, including those involving conditional image generation in latent diffusion models.

\section*{Acknowledgments}
\noindent
This work was partly supported by JSPS KAKENHI Grant Number 20H05952 and 23H00482.

%%%%%%%%% REFERENCES
{\small
\bibliographystyle{ieee_fullname}
\bibliography{egbib}

@article{ho2020denoising,
  title={Denoising diffusion probabilistic models},
  author={Ho, Jonathan and Jain, Ajay and Abbeel, Pieter},
  journal={Advances in neural information processing systems},
  volume={33},
  pages={6840--6851},
  year={2020}
}

@inproceedings{
song2021denoising,
title={Denoising Diffusion Implicit Models},
author={Jiaming Song and Chenlin Meng and Stefano Ermon},
booktitle={International Conference on Learning Representations},
year={2021}
}

@inproceedings{
meng2022sdedit,
title={{SDE}dit: Guided Image Synthesis and Editing with Stochastic Differential Equations},
author={Chenlin Meng and Yutong He and Yang Song and Jiaming Song and Jiajun Wu and Jun-Yan Zhu and Stefano Ermon},
booktitle={International Conference on Learning Representations},
year={2022}
}

@inproceedings{mokady2023null,
  title={Null-text inversion for editing real images using guided diffusion models},
  author={Mokady, Ron and Hertz, Amir and Aberman, Kfir and Pritch, Yael and Cohen-Or, Daniel},
  booktitle={Proceedings of the IEEE/CVF Conference on Computer Vision and Pattern Recognition},
  pages={6038--6047},
  year={2023}
}

@inproceedings{parmar2023zero,
  title={Zero-shot image-to-image translation},
  author={Parmar, Gaurav and Kumar Singh, Krishna and Zhang, Richard and Li, Yijun and Lu, Jingwan and Zhu, Jun-Yan},
  booktitle={ACM SIGGRAPH 2023 Conference Proceedings},
  pages={1--11},
  year={2023}
}

@inproceedings{pan2023effective,
  title={Effective real image editing with accelerated iterative diffusion inversion},
  author={Pan, Zhihong and Gherardi, Riccardo and Xie, Xiufeng and Huang, Stephen},
  booktitle={Proceedings of the IEEE/CVF International Conference on Computer Vision},
  pages={15912--15921},
  year={2023}
}

@article{garibi2024renoise,
  title={ReNoise: Real Image Inversion Through Iterative Noising},
  author={Garibi, Daniel and Patashnik, Or and Voynov, Andrey and Averbuch-Elor, Hadar and Cohen-Or, Daniel},
  journal={arXiv preprint arXiv:2403.14602},
  year={2024}
}

@article{white2016sampling,
  title={Sampling generative networks},
  author={White, Tom},
  journal={arXiv preprint arXiv:1609.04468},
  year={2016}
}

@inproceedings{avrahami2022blended,
  title={Blended diffusion for text-driven editing of natural images},
  author={Avrahami, Omri and Lischinski, Dani and Fried, Ohad},
  booktitle={Proceedings of the IEEE/CVF conference on computer vision and pattern recognition},
  pages={18208--18218},
  year={2022}
}

@inproceedings{
couairon2023diffedit,
title={DiffEdit: Diffusion-based semantic image editing with mask guidance},
author={Guillaume Couairon and Jakob Verbeek and Holger Schwenk and Matthieu Cord},
booktitle={The Eleventh International Conference on Learning Representations },
year={2023}
}

@inproceedings{
hertz2023prompttoprompt,
title={Prompt-to-Prompt Image Editing with Cross-Attention Control},
author={Amir Hertz and Ron Mokady and Jay Tenenbaum and Kfir Aberman and Yael Pritch and Daniel Cohen-or},
booktitle={The Eleventh International Conference on Learning Representations },
year={2023}
}

@inproceedings{rombach2022high,
  title={High-resolution image synthesis with latent diffusion models},
  author={Rombach, Robin and Blattmann, Andreas and Lorenz, Dominik and Esser, Patrick and Ommer, Bj{\"o}rn},
  booktitle={Proceedings of the IEEE/CVF conference on computer vision and pattern recognition},
  pages={10684--10695},
  year={2022}
}

@inproceedings{nichol2021improved,
  title={Improved denoising diffusion probabilistic models},
  author={Nichol, Alexander Quinn and Dhariwal, Prafulla},
  booktitle={International conference on machine learning},
  pages={8162--8171},
  year={2021},
  organization={PMLR}
}

@article{meiri2023fixed,
  title={Fixed-point Inversion for Text-to-image diffusion models},
  author={Meiri, Barak and Samuel, Dvir and Darshan, Nir and Chechik, Gal and Avidan, Shai and Ben-Ari, Rami},
  journal={arXiv preprint arXiv:2312.12540},
  year={2023}
}

@inproceedings{liu2015deep,
  title={Deep learning face attributes in the wild},
  author={Liu, Ziwei and Luo, Ping and Wang, Xiaogang and Tang, Xiaoou},
  booktitle={Proceedings of the IEEE international conference on computer vision},
  pages={3730--3738},
  year={2015}
}

@article{yu2015lsun,
  title={Lsun: Construction of a large-scale image dataset using deep learning with humans in the loop},
  author={Yu, Fisher and Seff, Ari and Zhang, Yinda and Song, Shuran and Funkhouser, Thomas and Xiao, Jianxiong},
  journal={arXiv preprint arXiv:1506.03365},
  year={2015}
}

@inproceedings{mahendran2015understanding,
  title={Understanding deep image representations by inverting them},
  author={Mahendran, Aravindh and Vedaldi, Andrea},
  booktitle={Proceedings of the IEEE conference on computer vision and pattern recognition},
  pages={5188--5196},
  year={2015}
}

@article{su2022dual,
  title={Dual diffusion implicit bridges for image-to-image translation},
  author={Su, Xuan and Song, Jiaming and Meng, Chenlin and Ermon, Stefano},
  journal={arXiv preprint arXiv:2203.08382},
  year={2022}
}

@inproceedings{kim2022diffusionclip,
  title={Diffusionclip: Text-guided diffusion models for robust image manipulation},
  author={Kim, Gwanghyun and Kwon, Taesung and Ye, Jong Chul},
  booktitle={Proceedings of the IEEE/CVF conference on computer vision and pattern recognition},
  pages={2426--2435},
  year={2022}
}
}

\end{document}